\documentclass{article} 
\usepackage{iclr2020_conference, times}

\usepackage{
    amsmath,
    amsfonts,
    bm
}

\def\eqref#1{equation~\ref{#1}}

\def\1{\bm{1}}

\def\vi{{\bm{i}}}
\def\vj{{\bm{j}}}
\def\vk{{\bm{k}}}

\def\mI{{\bm{I}}}

\def\mO{{\bm{O}}}

\def\mW{{\bm{W}}}

\DeclareMathAlphabet{\mathsfit}{\encodingdefault}{\sfdefault}{m}{sl}
\SetMathAlphabet{\mathsfit}{bold}{\encodingdefault}{\sfdefault}{bx}{n}

\def\gB{{\mathcal{B}}}
\def\gC{{\mathcal{C}}}

\def\gG{{\mathcal{G}}}

\def\gK{{\mathcal{K}}}

\def\gR{{\mathcal{R}}}

\def\sR{{\mathbb{R}}}
\def\sS{{\mathbb{S}}}

\def\sZ{{\mathbb{Z}}}

\usepackage{booktabs}
\usepackage{graphicx}
\usepackage{subcaption}

\usepackage{multirow}
\usepackage{makecell}

\usepackage{color}
\definecolor{dkgreen}{rgb}{0,0.6,0}
\definecolor{gray}{rgb}{0.5,0.5,0.5}
\definecolor{mauve}{rgb}{0.58,0,0.82}
\definecolor{calbrown}{RGB}{219,161,72}
\usepackage{verbatim}
\usepackage{hyperref}
\usepackage{lipsum}
\usepackage{xcolor}
\usepackage{relsize}
\usepackage{bbold}
\usepackage{url}

\usepackage{listings}
\newcommand{\bfsection}[1]{\noindent\textbf{#1:}}
\newcommand{\captiont}[2]{\caption{\textbf{#1} #2}}

\definecolor{large}{RGB}{0, 63, 245}
\definecolor{medium}{RGB}{136, 48, 142}
\definecolor{small}{RGB}{235, 64, 37}

\title{Deformable Kernels: Adapting Effective Receptive Fields for Object Deformation}

\author{
Hang Gao$^{1, 3}$\thanks{
        Equal contributions. Work is done when Hang and Xizhou are interns at
        Microsoft Research Asia.
    } ,
    Xizhou Zhu$^{2, 3}$\footnotemark[1] ,
    Steve Lin$^{3}$,
    Jifeng Dai$^{3}$ \\
    $^1$UC Berkeley
    $^2$University of Science and Technology of China
    $^3$Microsoft Research Asia \\
    \footnotesize{\texttt{hangg@eecs.berkeley.edu}, \texttt{ezra0408@mail.ustc.edu.cn}} \\
    \footnotesize{\texttt{\{stevelin,jifdai\}@microsoft.com}} \\
    \footnotesize{
        {
            \color{magenta}
            \url{http://people.eecs.berkeley.edu/\~hangg/deformable-kernels/}
        }
    }
}

\iclrfinalcopy  
\begin{document}

\maketitle

\begin{abstract}
Convolutional networks are not aware of an object's geometric variations, which leads to inefficient utilization of model and data capacity.
To overcome this issue, recent works on deformation modeling seek to spatially reconfigure the data towards a common arrangement such that semantic recognition suffers less from deformation.
This is typically done by augmenting static operators with learned free-form sampling grids in the image space, dynamically tuned to the data and task for adapting the receptive field.
Yet adapting the receptive field does not quite reach the actual goal~--~what really matters to the network is the \textit{effective} receptive field~(ERF), which reflects how much each pixel contributes.
It is thus natural to design other approaches to adapt the ERF directly during runtime.

In this work, we instantiate one possible solution as Deformable Kernels~(DKs), a family of novel and generic convolutional operators for handling object deformations by directly adapting the ERF while leaving the receptive field untouched.
At the heart of our method is the ability to resample the original kernel space towards recovering the deformation of objects.
This approach is justified with theoretical insights that the ERF is strictly determined by data sampling locations and kernel values.
We implement DKs as generic drop-in replacements of rigid kernels and conduct a series of empirical studies whose results conform with our theories.
Over several tasks and standard base models, our approach compares favorably against prior works that adapt during runtime.
In addition, further experiments suggest a working mechanism orthogonal and complementary to previous works.
\end{abstract}

\section{Introduction}

The rich diversity of object appearance in images arises from variations in object semantics and deformation.
Semantics describe the high-level abstraction of what we perceive, and
deformation defines the geometric transformation tied to specific
data~\citep{gibson1950perception}.
Humans are remarkably adept at making abstractions of the
world~\citep{hudson2019learning}; we see in raw visual signals, abstract
semantics away from deformation, and form concepts.

\begin{figure}[t]
    \centering
    \includegraphics[width=0.95\linewidth]{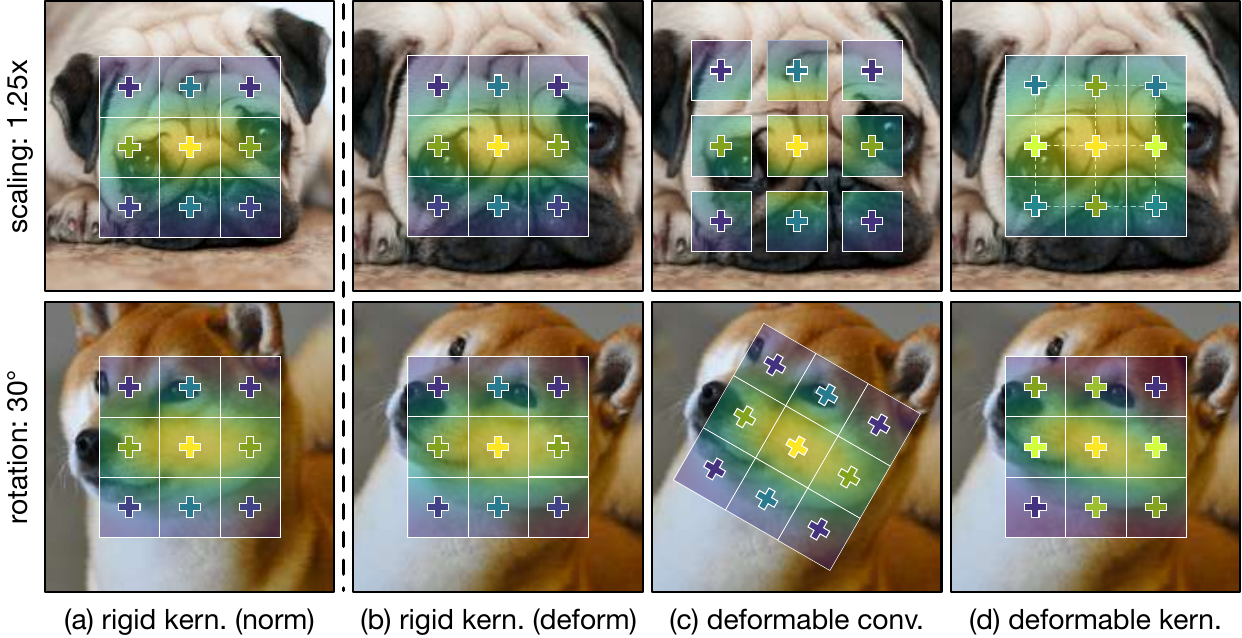}
    \captiont{Adaptation for deformation.}{
        We show how different $3\times3$ convolutions interact with deformations of two images.
        Kernel spaces are visualized as flat 2D Gaussians.
        Each ``$+$'' indicates a computation between a pixel and a kernel value
        sampled from the data and kernel space.
        Their colors represent corresponding kernel values.
        \textbf{(a, b)} Rigid kernels cannot adapt to specific deformations,
        thus consuming large model and data capacity.
        \textbf{(c)} Deformable Convolutions~\citep{dai2017deformable} reconfigure
        data towards common arrangement to counter the effects of geometric deformation.
        \textbf{(d)} Our Deformable Kernels~(DKs) instead resample kernels and,
        in effect, adapt kernel spaces while leaving the data untouched.
        Note that (b) and (c) share kernel values but sample different data
        locations, while (b) and (d) share data locations but sample different kernel values.
    }
    \vspace{-0.5em}
    \label{fig:teaser}
\end{figure}

Interestingly, modern convolutional networks follow an analogous process by
making abstractions through local connectivity and weight
sharing~\citep{zhang2019making}.  However, such a mechanism is an inefficient
one, as the emergent representations encode semantics and deformation together,
instead of as disjoint notions.  Though a convolution responds accordingly to
each input, how it responds is primarily programmed by its rigid kernels, as in
Figure~\hyperref[fig:teaser]{1(a, b)}.  In effect, this consumes large model
capacity and data modes~\citep{shelhamer2019blurring}.

We argue that the awareness of deformations emerges from adaptivity -- the
ability to adapt at
runtime~\citep{kanazawa2014locally,jia2016dynamic,li2019selective}.
Modeling of geometric transformations has been a constant
pursuit for vision researchers over
decades~\citep{lowe1999object,lazebnik2006beyond,jaderberg2015spatial,dai2017deformable}.
A basic idea is to spatially recompose data towards a common mode such that semantic
recognition suffers less from deformation.
A recent work that is representative of this direction is Deformable Convolution~\citep{dai2017deformable,zhu2019deformable}.
As shown in Figure~\hyperref[fig:teaser]{1(c)}, it augments the convolutions
with free-form sampling grids in the data space.
It is previously justified as adapting receptive field, or what we phrase as
the ``theoretical receptive field'', that defines \textit{which} input pixels
can contribute to the final output.
However, theoretical receptive field does not measure \textit{how much} impact
an input pixel actually has.
On the other hand, \citet{luo2016understanding} propose to measure the effective
receptive field~(ERF), i.e.\ the partial derivative of the output with respect to the
input data, to quantify the exact contribution of each raw pixel to the
convolution.
Since adapting the theoretical receptive field is not the goal but a means to adapt
the ERF, why not directly tune the ERF to specific data and tasks at runtime?

Toward this end, we introduce Deformable Kernels~(DKs), a family of novel and generic
convolutional operators for deformation modeling.
We aim to augment rigid kernels with the expressiveness to directly interact
with the ERF of the computation during inference.
Illustrated in Figure~\hyperref[fig:teaser]{1(d)}, DKs learn free-form offsets
on kernel coordinates to deform the original kernel space towards specific data
modality, rather than recomposing data.
This can directly adapt ERF while leaving receptive field untouched.
The design of DKs that is agnostic to data coordinates naturally
leads to two variants -- the global DK and the local DK, which behave
differently in practice as we later investigate.
We justify our approach with
theoretical results which show that ERF is strictly determined by data sampling locations and
kernel values.
Used as a generic drop-in replacement of rigid kernels, DKs achieve empirical
results coherent with our developed theory.
Concretely, we evaluate our operator with standard base models on
image classification and object detection.
DKs perform favorably against prior works that adapt during runtime.
With both quantitative and qualitative analysis, we further show that DKs can
work orthogonally and complementarily with previous techniques.

\section{Related Works}

We distinguish our work within the context of deformation modeling as our goal, and
dynamic inference as our means.

\bfsection{Deformation Modeling}
We refer to deformation modeling as learning geometric transformations in
2D image space without regard to 3D.
One angle to attack deformation modeling is to craft certain geometric
invariances into networks.
However, this usually requires designs specific to certain kinds of
deformation, such as shift, rotation, reflection and
scaling~\citep{sifre2013rotation,bruna2013invariant,kanazawa2014locally,cohen2016group,worrall2017harmonic,polar2018esteves}.
Another line of work on this topic learns to recompose data by either
semi-parameterized or completely free-form sampling in image space: Spatial
Transformers~\citep{jaderberg2015spatial} learns 2D affine transformations, Deep
Geometric Matchers~\citep{rocco2017convolutional} learns thin-plate spline
transformations, Deformable
Convolutions~\citep{dai2017deformable,zhu2019deformable} learns free-form
transformations.

We interpret sampling data space as an effective approach to adapt effective
receptive fields~(ERF) by directly changing receptive field.
At a high-level, our Deformable Kernels (DKs) share intuitions with this line
of works for learning geometric transformations, yet are instantiated by
learning to sample in kernel space which directly adapt ERF while leaving
theoretical receptive fields untouched.
While kernel space sampling is also studied in Deformable
Filter~\citep{xiong2019deformable} and KPConv~\citep{thomas2019kpconv}, but in
their contexts, sampling grids are computed from input point clouds rather than
learned from data corpora.

\bfsection{Dynamic Inference}
Dynamic inference adapts the model or individual operators to the observed
data.
The computation of our approach differs from
self-attention~\citep{vaswani2017attention,wang2018non} in which linear or
convolution modules are augmented with subsequent queries that extract from the same
input.
We consider our closest related works in terms of implementation as
those approaches that adapt convolutional kernels at run time.
It includes but is not limited to Dynamic Filters~\citep{jia2016dynamic},
Selective Kernels~\citep{li2019selective} and Conditional
Convolutions~\citep{yang2019soft}.
All of these approaches can learn and infer customized kernel spaces with respect to the
data, but are either less inefficient or are loosely formulated.
Dynamic Filters generate new filters from scratch, while Conditional
Convolutions extend this idea to linear combinations of a set of synthesized
filters.
Selective Kernels are, on the other hand, comparably lightweight, but
aggregating activations from kernels of different size is not as compact as
directly sampling the original kernel space.
Another line of works contemporary
to ours~\citep{shelhamer2019blurring,wang2019dynamic} is to compose free-form
filters with structured Gaussian filters, which essentially transforms kernel
spaces by data.
Our DKs also differ from these works with the emphasize of direct adaptation
the ERF rather than the theoretical receptive field.
As mentioned previously, the true goal should be to adapt the ERF, and to our knowledge, our work is the first to study dynamic inference of ERFs.

\section{Approach}

We start by covering preliminaries on convolutions, including the
definition of effective receptive field~(ERF).
We then formulate a theoretical framework for analyzing ERFs, and thus motivate
our Deformable Kernels~(DKs).
We finally elaborate different DK variants within such a framework.
Our analysis suggests compatibility between DKs and the prior work.

\subsection{A Dive into Convolutions} \label{sec:diving}

\bfsection{2D Convolution}
Let us first consider an input image~$\mI \in \sR^{D \times D}$.
By convolving it with a kernel~$\mW \in \sR^{K \times K}$ of stride~$1$, we
have an output image~$\mO$ whose pixel values at each coordinate~$\vj \in
\sR^2$ can be expressed as
\begin{equation}
\small
    \label{eq:conv2d}
    \mO_{\vj} =
    \sum_{\vk \in \gK} \mI_{\vj + \vk} \mW_{\vk},
\end{equation}
by enumerating discrete kernel positions $\vk$ within the support~$\gK = [-K /
2, K / 2]^2 \cap \sZ$. This defines a rigid grid for sampling data and kernels.

\bfsection{Theoretical Receptive Field}
The same kernel~$\mW$ can be stacked repeatedly to
form a \textit{linear} convolutional network with~$n$ layers.
The theoretical receptive field can then be imagined as the ``accumulative
coverage'' of kernels at each given output unit on the input image by
deconvolving back through the network.
This property characterizes a set of input fields that could fire
percepts onto corresponding output pixels.
The size of a theoretical receptive field scales linearly with respect to the
network depth~$n$ and kernel size~$K$~\citep{he2016deep}.

\bfsection{Effective Receptive Field}
Intuitively, not all pixels within a theoretical receptive field contribute
equally.
The influence of different fields varies from region to region
thanks to the central emphasis of stacked convolutions and also to the non-linearity induced
by activations.
The notion of \textit{effective} receptive field
(ERF)~\citep{luo2016understanding} is thus introduced to measure the
impact of each input pixel on the output at given locations.
It is defined as a partial derivative field of the output with respect to the input
data.
With the numerical approximations in linear convolution networks, the ERF was
previously identified as a Gaussian-like soft attention map over input images
whose size grows \textit{fractionally} with respect to the network depth~$n$ and linearly to the kernel size~$K$.
Empirical results validate this idea under more complex and
realistic cases when networks exploit non-linearities, striding, padding, skip
connections, and subsampling.

\subsection{Analysis on Effective Receptive Fields} \label{sec:analyzing}

We aim to revisit and complement the previous analysis on ERFs
by~\citet{luo2016understanding}.
While the previous analysis concentrates on studying the expectation of an ERF,
i.e., when network depth~$n$ approaches infinity or all kernels are randomly
distributed without learning in general, our analysis focuses on how we can
perturb the computation such that the change in ERF is predictable,
given an input and a set of kernel spaces.\

We start our analysis by considering a \textit{linear} convolutional
network, without any unit activations, as defined in Section~\ref{sec:diving}.
For consistency, superscripts are introduced to image~$\mI$, kernel~$\mW$, and
subscripts to kernel positions~$\vk$ to denote the index $s \in [1, n]$ of each
layer.
Formally, given an input image~$\mI^{(0)}$ and a set of~$K \times K$
kernels~$\{\mW^{(s)}\}_{s = 1}^{n}$ of stride~$1$, we can roll out the final
output $\mO \equiv \mI^{(n)}$ by unfolding Equation~\ref{eq:conv2d} as
\begin{equation}
\small
    \label{eq:unfolding}
    \begin{split}
        \mI^{(n)}_{\vj}
        &=
        \sum_{\vk_{n} \in \gK}
        \mI_{\vj + \vk_{n}}^{(n - 1)}
        \mW_{\vk_{n}}^{(n)}
        =
        \sum_{(\vk_{n - 1}, \vk_{n}) \in \gK^2}
        \mI_{\vj + \vk_{n} + \vk_{n - 1}}^{(n - 2)}
        \mW_{\vk_{n}}^{(n)}
        \mW_{\vk_{n - 1}}^{(n - 1)}
        =
        \cdots
        \\
        &=
        \mathlarger{\mathlarger{\sum}}_{(\vk_1, \vk_2, \ldots, \vk_n) \in \gK^n}
        \left(
            \mI_{\vj + \sum_{s = 1}^{n} \vk_{s}}^{(0)}
            \cdot
            \prod_{s = 1}^{n} \mW^{(s)}_{\vk_s}
        \right).
    \end{split}
\end{equation}
By definition\footnote{\
    The original definition of ERF in
    \cite{luo2016understanding} focuses on the central coordinate of the output,
    i.e.~$\vj = (0, 0)$, to partially avoid the effects of zero padding.
    In this work, we will keep $\vj$ in favor of generality while explicitly
    assuming input size~$D \to \infty$.
}, the effective receptive field value~$\gR^{(n)}(\vi; \vj) \equiv \partial
\mI^{(n)}_{\vj} / \partial \mI^{(0)}_{\vi}$ of output coordinate~$\vj$ that
takes input coordinate~$\vi$ can be computed by
\begin{equation}
\small
    \label{eq:erf-grad}
    \gR^{(n)}(\vi; \vj)
    =
    \mathlarger{\mathlarger{\sum}}_{(\vk_1, \vk_2, \ldots, \vk_n) \in \gK^n}
    \left(
        \mathbb{1}\big[
            \vj + \sum_{s = 1}^{n} \vk_{s} = \vi
        \big]
        \cdot
        \prod_{s = 1}^{n} \mW^{(s)}_{\vk_s}
    \right),
\end{equation}
where~$\mathbb{1}[\cdot]$ denotes the indicator function. This result indicates
that ERF is related only to the data sampling location~$\vj$, kernel sampling
location~$\vk$, and kernel matrices~$\{\mW^{(s)}\}$.

If we replace the~$m^{\text{th}}$ kernel~$\mW^{(m)}$ with a~$1 \times 1$
kernel of a single parameter~$\mW^{(m)}_{\tilde{\vk}_m}$ sampled from it, the
value of ERF becomes to
\begin{equation}
\small
    \label{eq:erf-grad-m}
    \gR^{(n)}(\vi; \vj, \tilde{\vk}_m)
    =
    \mathlarger{\mathlarger{\sum}}_{\
        (\vk_1, \ldots, \vk_{m-1}, \vk_{m+1}, \ldots, \vk_{n}) \in \gK^{n-1}
    }
    \left(
        \mathbb{1}\big[
            \vj + \sum_{s \in \sS} \vk_{s} = \vi
        \big]
        \cdot
        \prod_{s \in \sS} \mW^{(s)}_{\vk_s}
        \cdot
        \mW^{(m)}_{\tilde{\vk}_m}
    \right),
\end{equation}
where~$\sS = [1, n] \setminus \{m\}$. Since a~$K \times K$ kernel can be deemed
as a composition of~$K^2$ $1 \times 1$ kernels distributed on a square
grid, Equation~\ref{eq:erf-grad} can thus be reformulated as
\begin{equation}
\small
    \label{eq:erf-grad-reformulate}
    \gR^{(n)}(\vi; \vj)
    =
    \sum_{\vk_{m} \in \gK}
    \gR^{(n)}(\vi; \vj + \vk_m, \vk_m).
\end{equation}
For the case of complex non-linearities, where we here consider post ReLU\footnote{\
    Our analysis currently only considers the ReLU network for its
    nice properties and prevalent popularity.
} activations in
Equation~\ref{eq:conv2d},
\begin{equation}
\small
    \label{eq:conv2d-relu}
    \mO_{\vj} =
    \max(\sum_{\vk \in \gK} \mI_{\vj + \vk} \mW_{\vk}, 0).
\end{equation}
We can follow a similar analysis and derive corresponding ERF as
\begin{equation*}
\small
\label{eq:erf-grad-m-relu}
\begin{split}
    &
    \gR^{'(n)}(\vi; \vj, \tilde{\vk}_m)
    =
    \sum_{
        (\vk_1, \cdots, \vk_{m-1}, \vk_{m+1}, \cdots, \vk_n)
        \in
        \gK^{n - 1}
    }
    \left(
        \gC^{(n)}(\vi; \vj, \vk_1, \cdots, \vk_n, \tilde{\vk}_m)
        \cdot
        \prod_{s \in \sS} \mW_{\vk_s}^{(s)}
        \cdot
        \mW_{\tilde{\vk}_m}^{m}
    \right)
    \\
    \text{where ~}
    &
    \gC^{(n)}(\vi; \vj, \vk_1, \cdots, \vk_n, \tilde{\vk}_m)
    =
    \mathbb{1}\big[
        \vj + \sum_{s \in \sS} \vk_s = \vi
    \big]
    \prod_{s \in \sS}
    \mathbb{1}\big[
        \mI_{\vj}^{(s - 1)} \mW_{\tilde{\vk}_s}^{(s)} > 0
    \big]
    \mathbb{1}\big[
        \mI_{\vj}^{(m - 1)} \mW_{\tilde{\vk}_m}^{(m)} > 0
    \big].
\end{split}
\end{equation*}
Here we can see that the ERF becomes data-dependent due to the coefficient~$\gC$,
which is tied to input coordinates, kernel sampling locations, and input
data~$\mI^{(0)}$.
The more detailed analysis of this coefficient is beyond the scope of
this paper.
However, it should be noted that this coefficient only ``gates'' the
contribution of the input pixels to the output.
So in practice, ERF is ``porous'' -- there are inactive (or gated) pixel units
irregularly distributed around the ones that fire.
This phenomenon also appeared in previous studies (such as in
\citet{luo2016understanding}, Figure 1).
The maximal size of an ERF is still controlled by the data sampling location and
kernel values as in the linear cases in Equation~\ref{eq:erf-grad-reformulate}.

A nice property of Equation~\ref{eq:erf-grad-m} and
Equation~\ref{eq:erf-grad-reformulate} is that all computations are
linear, making it compatible with any linear sampling operators for querying
kernel values of fractional coordinates.
In other words, sampling kernels in effect samples the ERF on the data in the
linear case, but also roughly generalizes to non-linear cases as well.
This finding motivates our design of Deformable Kernels~(DKs) in
Section~\ref{sec:dk}.

\subsection{Deformable Kernels} \label{sec:dk}

In the context of Equation~\ref{eq:conv2d}, we resample the kernel
$\mW$ with a group of \textit{learned} kernel offsets denoted as $\{\Delta
\vk\}$ that correspond to each discrete kernel position $\vk$.
This defines our DK as
\begin{equation}
\small
    \label{eq:dk2d}
    \mO_{\vj} =
    \sum_{\vk \in \gK} \mI_{\vj + \vk} \mW_{\vk + \Delta \vk},
\end{equation}
and the value of ERF as
\begin{equation}
\small
    \label{eq:erf-grad-dk}
    \gR^{(n)}_{\text{DK}}(\vi; \vj)
    =
    \sum_{\vk_{m} \in \gK}
    \gR^{(n)}(\vi; \vj + \vk_m, \vk_m + \Delta \vk_m).
\end{equation}
Note that this operation leads to sub-pixel sampling in the kernel space.
In practice, we use bilinear sampling to interpolate within the discrete kernel
grid.

Intuitively, the size (resolution) of the original kernel space can
affect sampling performance.
Concretely, suppose we want to sample a~$3 \times 3$ kernel.
DKs do not have any constraint on the size of the original kernel space, which
we call the ``scope size'' of DKs.
That said, we can use a~$\mW$ of any size~$K'$ even though the number of
sampling locations is fixed as~$K^2$.
We can thus exploit large kernels -- the largest ones can reach~$9 \times 9$ in
our experiments with nearly no overhead in computation since bilinear
interpolations are extremely lightweight compared to the cost of
convolutions.
This can also increase the number of learning parameters, which in practice
might become intractable if not handled properly.
In our implementation, we will exploit depthwise
convolutions~\citep{howard2017mobilenets} such that increasing scope size
induces a negligible amount of extra parameters.

\begin{figure}[t]
    \centering
    \includegraphics[width=0.95\linewidth]{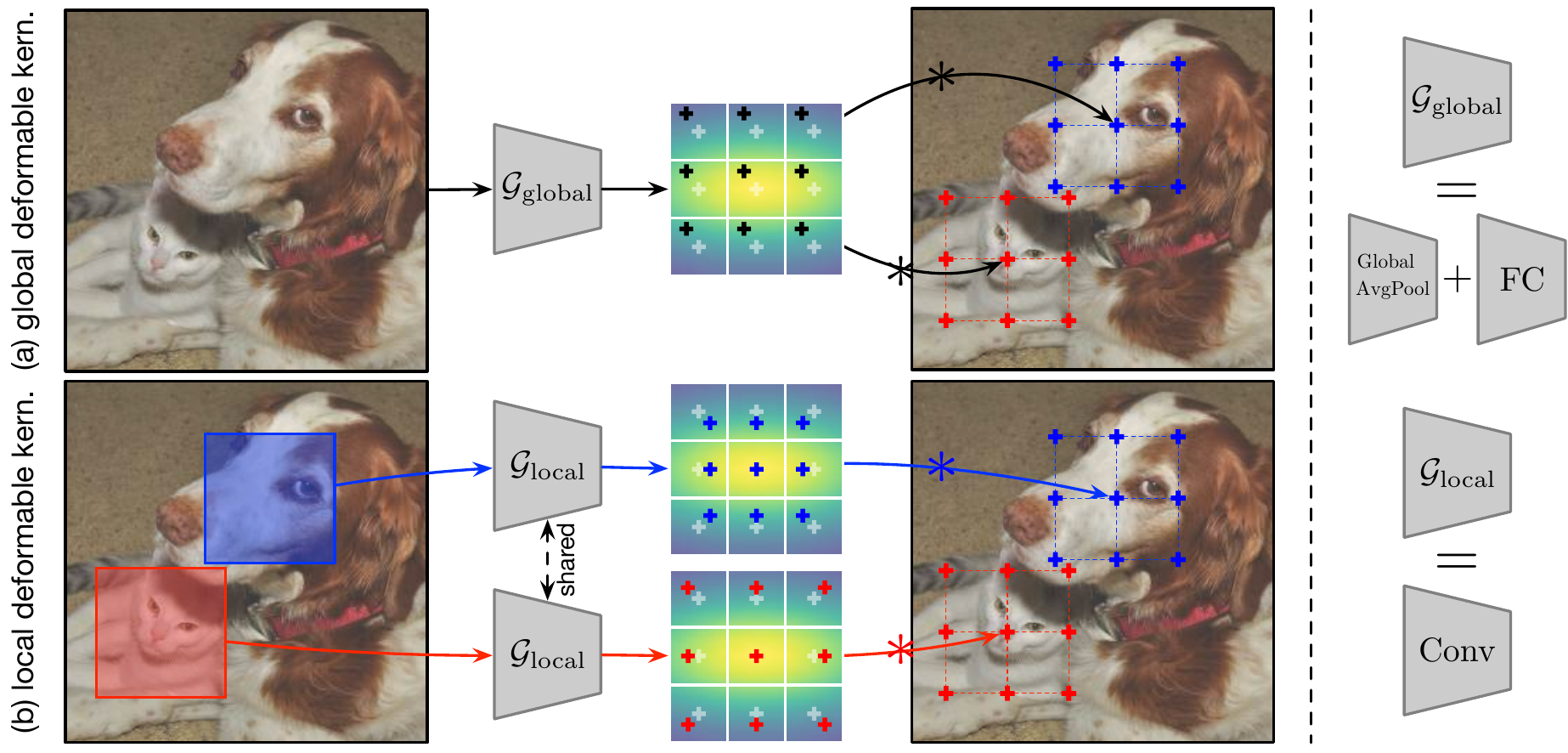}
    \captiont{Instantiations.}{\
        We show how DK variants works with an example image that contains a
        \textcolor{large}{large} and a \textcolor{small}{small} object.
        \textbf{(a)} The global DK learns one set of kernel sampling grid given
        an input image and apply it to all data positions.
        \textbf{(b)} The local DK adapts kernels for each input patches, and
        induces better locality for deformation modeling.
    }%
    \vspace{-0.5em}
    \label{fig:operators}
\end{figure}

As previously discussed, sampling the kernel space in effect transforms into
sampling the ERF.\
On the design of locality and spatial granularity of our learned offsets,
DK naturally delivers two variants -- the global DK and the local DKs, as
illustrated in Figure~\ref{fig:operators}.
In both operators, we learn a kernel offset generator~$\gG$ that maps an input
patch into a set of kernel offsets that are later applied to rigid kernels.

In practice, we implement $\gG_{\text{global}}$ as a stack of one global
average pooling layer, which reduces feature maps into a vector, and another
fully-connected layer without non-linearities, which projects the reduced
vector into an offset vector of $2 K^2$ dimensions.
Then, we apply these offsets to \textit{all} convolutions for the input image
following Equation~\ref{eq:dk2d}.
For local DKs, we implement $\gG_{\text{local}}$ as an extra convolution that
has the same configuration as the target kernel, except that it only has $2 K^2$
output channels.
This produces kernel sampling offsets $\{\Delta \vk\}$ that are additionally
indexed by output locations $\vj$.
It should be noted that similar designs were also discussed
in~\citet{jia2016dynamic}, in which filters are generated given either an image
or individual patches from scratch rather than by resampling.

Intuitively, we expect the global DK to adapt kernel space between different
images but not within a single input.
The local DK can further adpat to specific image patches: for smaller objects,
it is better to have shaper kernels and thus denser ERF; for larger objects,
flatter kernels can be more beneficial for accumulating a wider ERF.\
On a high level, local DKs can preserve better locality and have larger freedom
to adapt kernel spaces comparing to its global counterpart.
We later compare these operators in our experiments.

\subsection{Link with Deformable Convolutions} \label{sec:dcn}
The core idea of DKs is to learn adaptive offsets to sample the kernel space for
modeling deformation, which makes them similar to Deformable
Convolutions~\citep{dai2017deformable,zhu2019deformable}, at both the conceptual and
implementation levels.
Here, we distinguish DKs from Deformable Convolutions and show how they can be unified.

Deformable Convolutions can be reformulated in a general form as
\begin{equation}
\small
    \label{eq:dc2d}
    \mO_{\vj} =
    \sum_{\vk \in \gK} \mI_{\vj + \vk + \Delta \vj} \mW_{\vk},
\end{equation}
where they aim to learn a group of data offsets~$\{\Delta \vj\}$
with respect to discrete data positions~$\vj$.
For consistency for analysis, the value of effective receptive field becomes
\begin{equation}
\small
    \label{eq:erf-grad-dc}
    \gR^{(n)}_{\text{DC}}(\vi; \vj)
    =
    \sum_{\vk_{m} \in \gK}
    \gR^{(n)}(\vi; \vj + \vk_m + \Delta \vj_m, \vk_m).
\end{equation}
This approach essentially recomposes the input image towards common modes such that
semantic recognition suffers less from deformation.
Moreover, according to our previous analysis in
Equation~\ref{eq:erf-grad-reformulate}, sampling data is another way of
sampling the ERF.\
This, to a certain extent, also explains why Deformable Convolutions are well
suited for learning deformation-agnostic representations.

Moreover, we can learn both data \textit{and} kernel offsets in one
convolutional operator.
Conceptually, this can be done by merging Equation~\ref{eq:dk2d} with
Equation~\ref{eq:dc2d}, which leads to
\begin{equation}
\small
    \label{eq:dkdc2d}
    \begin{aligned}
        \mO_{\vj}
        &=
        \sum_{\vk \in \gK}
        \mI_{\vj + \vk + \Delta \vj} \mW_{\vk + \Delta \vk},
        \\
        \gR^{(n)}_{\text{DC+DK}}(\vi; \vj)
        &=
        \sum_{\vk_{m} \in \gK}
        \gR^{(n)}(\vi; \vj + \vk_m + \Delta \vj_m, \vk_m + \Delta \vk_m).
    \end{aligned}
\end{equation}
We also investigate this operator in our experiments.
Although the two techniques may be viewed as serving a similar
purpose, we find the collaboration between Deformable
Kernels and Deformable Convolutions to be powerful in practice, suggesting strong compatibility.

\section{Experiments}

We evaluate our Deformable Kernels~(DKs) on image classification using ILSVRC
and object detection using the COCO benchmark.
Necessary details are provided to reproduce our results, together with
descriptions on base models and strong baselines for all experiments and
ablations.
For task-specific considerations, we refer to each corresponding section.

\bfsection{Implementation Details}
We implement our operators in PyTorch and CUDA.\
We exploit depthwise convolutions when designing our operator for better
computational efficiency\footnote{\
    This makes enlarging the kernel scope size tractable and prevents extensive
    resource competition in CUDA kernels when applying local DKs.
}.
We initialize kernel grids to be uniformly distributed within the scope size.
For the kernel offset generator, we set its learning rate to be a fraction of
that of the main network, which we cross-validate for each base model.
We also find it important to clip sampling locations inside the original kernel
space, such that~$\vk + \Delta \vk \in \gK$ in Equation~\ref{eq:dk2d}.

\bfsection{Base Models}
We choose our base models to be ResNet-50~\citep{he2016deep} and
MobileNet-V2~\citep{sandler2018mobilenetv2}, following the standard practice
for most vision applications.
As mentioned, we exploit depthwise convolution and thus make changes to the
ResNet model.
Concretely, we define our \textit{ResNet-50-DW} base model by replacing all~$3
\times 3$ convolutions by its depthwise counterpart while doubling the
dimension of intermediate channels in all residual blocks.
We find it to be a reasonable base model compared to the original ResNet-50,
with comparable performance on both tasks.
During training, we set the weight decay to be~$4 \times 10^{-5}$ rather than
the common~$10^{-4}$ for both models since depthwise models usually underfit rather
than overfit~\citep{xie2017aggregated,howard2017mobilenets,hu2018squeeze}.
We set the learning rate multiplier of DK operators as~$10^{-2}$ for
ResNet-50-DW and~$10^{-1}$ for MobileNet-V2 in all of our experiments.

\bfsection{Strong Baselines}
We develop our comparison with two previous works: Conditional
Convolutions~\citep{yang2019soft} for dynamics inference, and Deformable
Convolutions~\citep{dai2017deformable,zhu2019deformable} for deformation
modeling.
We choose Conditional Convolutions due to similar computation forms --
sampling can be deemed as an elementewise ``expert voting'' mechanism.
For fair comparisons, We reimplement and reproduce their results.
We also combine our operator with these previous approach to show
both quantitative evidence and qualitative insight that our working mechanisms
are compatible.

\subsection{Image Classification} \label{sec:cls}

We first train our networks on the ImageNet 2012 training
set~\citep{deng2009imagenet}.
We adopt a common experiment protocol for fair comparisons as in~\citet{goyal2017accurate,loshchilov2016sgdr}.
For more details, please refer to our supplement.

We first ablate the scope size of kernels for our DKs and study how it can
affect model performance using ResNet-50-DW.
As shown in Table~\ref{tab:cls_ablation_space}, our DKs are sensitive to the
choice of the scope size.
We shown that when only applied to $3 \times 3$ convolutions inside residual
bottlenecks, local DKs induce a +0.7 performance gain within the original
scope.
By further enlarging the scope size, performance increases yet quickly plateaus
at scope $4 \times 4$, yielding largest +1.4 gain for top-1 accuracy.
Our speculation is that, although increasing scope size theoretically means
better interpolation, it also makes the optimization space exponentially larger
for each convolutional layer.
And since number of entries for updating is fixed, this also leads to
relatively sparse gradient flows.
In principle, we set default scope size at $4 \times 4$ for our DKs.

\setlength{\tabcolsep}{5pt}
\renewcommand{\arraystretch}{1.1}
\begin{table}[t]
    \small
    \centering
    \resizebox{0.95\linewidth}{!}{
        \begin{tabular}{c|c|c|ccc}
            \Xhline{1.0pt}
            Backbone & 1$\times$1 Deformable Kernels & \multicolumn{1}{c|}{3$\times$3 Deformable Kernels} & top1 (\%) & \#P (M) & GFLOPs \\
            \hline
            \hline
            ResNet-50 & w/o & w/o & 76.7 & 25.6 & 3.86 \\
            \hline
            ResNet-50-DW & w/o & w/o & 76.3 & 23.7 & 3.82 \\
            \hline
            \multirow{4}{*}{ResNet-50-DW} & \multirow{4}{*}{w/o} & local, scope size 3$\times$3 & 77.4 & 24.9 & 4.32 \\
            \cline{3-6}
             & & \bfseries{local, scope size 4$\times$4} & \bfseries{78.1} & \bfseries{25.0} & \bfseries{4.32} \\
            \cline{3-6}
             & & local, scope size 5$\times$5 & 77.8 & 25.0 & 4.32 \\
            \cline{3-6}
             & & local, scope size 9$\times$9 & 77.4 & 25.4 & 4.32 \\
            \hline
            \hline
            \multirow{4}{*}{ResNet-50-DW} & w/o & global, scope size 4$\times$4 & 77.6 & 23.9 & 3.82 \\
            \cline{2-6}
             & global, scope size 2$\times$2 & global, scope size 4$\times$4 & 77.9 & 80.1 & 4.09 \\
            \cline{2-6}
             & w/o & local, scope size 4$\times$4 & 78.1 & 25.0 & 4.32 \\
            \cline{2-6}
             & \bfseries{global, scope size 2$\times$2} & \bfseries{local, scope size 4$\times$4} & \bfseries{78.5} & \bfseries{81.2} & \bfseries{4.60} \\
            \hline
            \hline
            MobileNet-V2 & w/o & w/o & 71.9 & 3.5 & 0.31 \\
            \hline
            \multirow{4}{*}{MobileNet-V2} & w/o & global, scope size 4$\times$4 & 73.6 & 3.7 & 0.31 \\
            \cline{2-6}
             & global, scope size 2$\times$2 & global, scope size 4$\times$4 & 74.5 & 10.1 & 0.34 \\
            \cline{2-6}
             & w/o & local, scope size 4$\times$4 & 74.1 & 4.7 & 0.73 \\
            \cline{2-6}
             & \bfseries{global, scope size 2$\times$2} & \bfseries{local, scope size 4$\times$4} & \bfseries{74.8} & \bfseries{11.1} & \bfseries{0.76} \\
             \Xhline{1.0pt}
         \end{tabular}
    }
    \captiont{Ablations of scope size and different instantiations of DK for
        image classification.}{
        Using proper scope size, and more DK layers boosts performance.
        Modeling individual offset kernel grid for each data entries is also
        beneficial.
    }
    \vspace{-0.5em}
    \label{tab:cls_ablation_space}
\end{table}

\setlength{\tabcolsep}{5pt}
\renewcommand{\arraystretch}{1.1}
\begin{table}[t]
    \small
    \centering
    \resizebox{0.95\linewidth}{!}{
        \begin{tabular}{c|c|c|ccc}
            \Xhline{1.0pt}
            Backbone & 1$\times$1 Deformable Kernels & \multicolumn{1}{c|}{3$\times$3 Deformable Kernels} & top1 (\%) & \#P (M) & GFLOPs \\
            \hline
            \hline
            ResNet-50-DW & w/o & local, scope size 4$\times$4 & 78.1 & 25.0 & 4.32 \\
            \hline
             \multirow{2}{*}{\makecell{ResNet-50-DW\\with SCC}} & \multirow{2}{*}{w/o} & w/o & 77.6 & 42.5 & 7.13 \\
             \cline{3-6}
             & & local, scope size 4$\times$4 & 78.9 & 43.7 & 7.61 \\
             \hline
             \multirow{2}{*}{\makecell{ResNet-50-DW\\with DCN}} & \multirow{2}{*}{w/o} & w/o & 78.0 & 24.8 & 4.10\\
             \cline{3-6}
             & & \textbf{local, scope size 4$\times$4} & \textbf{79.0} & \textbf{26.1} & \textbf{4.60} \\
             \hline
             \hline
            MobileNet-V2 & w/o & local, scope size 4$\times$4 & 74.1 & 4.7 & 0.73 \\
             \hline
             \multirow{2}{*}{\makecell{MobileNet-V2\\with SCC}} & \multirow{2}{*}{w/o} & w/o & 74.3 & 19.0 & 2.19 \\
             \cline{3-6}
             & & \textbf{local, scope size 4$\times$4} & \textbf{75.5} & \textbf{19.7} & \textbf{2.48} \\
             \hline
             \multirow{2}{*}{\makecell{MobileNet-V2\\with DCN}} & \multirow{2}{*}{w/o} & w/o & 73.2 & 4.6 & 0.52 \\
             \cline{3-6}
             & & local, scope size 4$\times$4 & 74.4 & 5.8 & 0.93 \\
             \Xhline{1.0pt}
         \end{tabular}
    }
    \captiont{Comparisons to strong baselines for image classification}{
        DKs perform comparably or superiorly to previous methods.
        Further combinations yield consistent gain, suggesting orthogonal and
        compatible working mechanisms.
    }
    \vspace{-0.5em}
    \label{tab:cls_combination}
\end{table}

We next move on and ablate our designs by comparing the global DK with the
local DK, shown in the table.
Both operators helps while the local variants consistently performs better than
their global counterparts, bringing a +0.5 gap on both base models.
We also study the effect of using more DKs in the models -- the $1 \times 1$
convolutions are replaced by global DKs\footnote{\
    The implementation of local DKs right now cannot support large number of
    output channels.
} with scope $2 \times 2$.
Note that all $1 \times 1$ convolutions are not depthwise, and therefore
this operation induces nearly 4 times of parameters.
We refer their results only for ablation and show that adding more DKs
still helps -- especially for MobileNet-V2 since it is under-parameterized.
This finding also holds for previous models~\citep{yang2019soft} as well.

We further compare and combine DKs with Conditional Convolutions and Deformable
Convolutions.
Results are recorded in Table~\ref{tab:cls_combination}.
We can see that DKs perform comparably on ResNet-V2 and compare favorably on
MobileNet-V2 -- improve +0.9 from Deformable Convolutions and achieve comparable
results with less than a quarter number of parameters compared to Conditional
Convolutions.
Remarkably, we also show that if combined together, even larger performance
gains are in reach.
We see consistent boost in top-1 accuracy compared to strong baselines:
+1.3/+1.0 on ResNet-50-DW, and +1.2/+1.2 on MobileNet-V2.
These gaps are bigger than those from our own ablation, suggesting the working
mechanisms across the operators to be orthogonal and compatible.

\subsection{Object Detection} \label{sec:det}

\setlength{\tabcolsep}{5pt}
\renewcommand{\arraystretch}{1.1}
\begin{table}[t]
    \small
    \centering
    \resizebox{0.95\linewidth}{!}{
        \begin{tabular}{c|c|c|cccc}
            \Xhline{1.0pt}
            Backbone & 1$\times$1 Deformable Kernels & \multicolumn{1}{c|}{3$\times$3 Deformable Kernels} & mAP & mAP$_\text{S}$ & mAP$_\text{M}$ & mAP$_\text{L}$ \\
            \hline
            \hline
            ResNet-50-DW & w/o & w/o & 36.6 & 22.1 & 39.9 & 46.6 \\
            \hline
            \multirow{4}{*}{ResNet-50-DW} & w/o & global, scope size 4$\times$4 & 36.7 & 22.6 & 40.2 & 46.9\\
            \cline{2-7}
             & global, scope size 2$\times$2 & global, scope size 4$\times$4 & 37.1 & 23.1 & 40.6 & 46.6 \\
            \cline{2-7}
             & w/o & local, scope size 4$\times$4 & 37.8 & 23.4 & 41.6 & 48.2 \\
            \cline{2-7}
             & \bfseries{global, scope size 2$\times$2} & \bfseries{local, scope size 4$\times$4} & \bfseries{38.4} & \bfseries{23.4} & \bfseries{42.0} & \bfseries{49.4} \\
            \hline
            \hline
            MobileNet-V2 & w/o & w/o & 31.3 & 18.6 & 33.7 & 40.4 \\
            \hline
            \multirow{4}{*}{MobileNet-V2} & w/o & global, scope size 4$\times$4 & 32.5 & 19.6 & 35.8 & 41.9 \\
            \cline{2-7}
             & global, scope size 2$\times$2 & global, scope size 4$\times$4 & 32.9 & 19.4 & 35.5 & 42.8 \\
            \cline{2-7}
             & w/o & local, scope size 4$\times$4 & 32.9 & 19.5 & 36.0 & 42.5 \\
            \cline{2-7}
             & \bfseries{global, scope size 2$\times$2} & \bfseries{local, scope size 4$\times$4} & \bfseries{33.7} & \bfseries{20.2} & \bfseries{36.7} & \bfseries{44.0} \\
             \Xhline{1.0pt}
         \end{tabular}
    }
    \captiont{Ablations for object detection.}{
        Consistent results with image classification.
    }
    \vspace{-0.5em}
    \label{tab:det_ablation_patchwise}
\end{table}

\setlength{\tabcolsep}{5pt}
\renewcommand{\arraystretch}{1.1}
\begin{table}[t]
    \small
    \centering
    \resizebox{0.95\linewidth}{!}{
        \begin{tabular}{c|c|c|cccc}
            \Xhline{1.0pt}
            Backbone & 1$\times$1 Deformable Kernels & \multicolumn{1}{c|}{3$\times$3 Deformable Kernels} & mAP & mAP$_\text{S}$ & mAP$_\text{M}$ & mAP$_\text{L}$ \\
            \hline
            \hline
            ResNet-50-DW & global, scope size 2$\times$2 & local, scope size 4$\times$4 & 38.4 & 23.4 & 42.0 & 49.4 \\
            \hline
             \multirow{2}{*}{\makecell{ResNet-50-DW\\with SCC}} & \multirow{2}{*}{w/o} & w/o & 36.3 & 22.1 & 39.3 & 47.0 \\
             \cline{3-7}
             & & local, scope size 4$\times$4 & 38.0 & 23.4 & 41.9 & 48.4 \\
             \hline
             \multirow{2}{*}{\makecell{ResNet-50-DW\\with DCN}} & \multirow{2}{*}{w/o} & w/o & 39.9 & 24.0 & 43.4 & 52.6 \\
             \cline{3-7}
             & & \textbf{local, scope size 4$\times$4} & \textbf{40.6} & \textbf{24.6} & \textbf{43.9} & \textbf{53.3} \\
             \hline
             \hline
            MobileNet-V2 & global, scope size 2$\times$2 & local, scope size 4$\times$4 & 33.7 & 20.2 & 36.7 & 44.0 \\
            \hline
             \multirow{2}{*}{\makecell{MobileNet-V2\\with SCC}} & \multirow{2}{*}{w/o} & w/o & 33.2 & 20.5 & 35.6 & 43.3 \\
             \cline{3-7}
             & & local, scope size 4$\times$4 & 34.3 & 20.2 & 37.3 & 44.7 \\
             \hline
             \multirow{2}{*}{\makecell{MobileNet-V2\\with DCN}} & \multirow{2}{*}{w/o} & w/o & 34.4 & 20.5 & 37.0 & 44.7 \\
             \cline{3-7}
             & & \textbf{local, scope size 4$\times$4} & \textbf{35.6} & \textbf{20.6} & \textbf{38.5} & \textbf{47.3} \\
             \Xhline{1.0pt}
         \end{tabular}
    }
    \captiont{Comparisons to strong baselines for object detection}{
        DKs perform fall short to Deformable Convolution, but combination still
        improves performance.
    }
    \vspace{-0.5em}
    \label{tab:det_combination}
\end{table}

We examine DKs on the COCO benchmark~\citep{lin2014microsoft}.
For all experiments, we use Faster R-CNN~\citep{ren2015faster} with
FPN~\citep{lin2017feature} as the base detector, plugging in the backbones we
previously trained on ImageNet.
For MobileNet-V2, we last feature maps of the each resolution for FPN post
aggregation.
Following the standard protocol, training and evaluation are performed on the
$120$k images in the \texttt{train-val} split and the $20$k images in the
\texttt{test-dev} split, respectively.
For evaluation, we measure the standard mean average precision (mAP) and
shattered scores for small, medium and large objects.

Table~\ref{tab:det_ablation_patchwise} and Table~\ref{tab:det_combination}
follow the same style of analysis as in image classification.
While the baseline methods of ResNet achieve 36.6 mAP, indicating a strong
baseline detector, applying local DKs brings a +1.2 mAP
improvement when replacing 3x3 rigid kernels alone and a +1.8 mAP improvement
when replacing both 1x1 and 3x3 rigid kernels.
This trend magnifies on MobileNet-v2 models, where we see an improvement of
+1.6 mAP and +2.4 mAP, respectively.
Results also confirm the effectiveness of local DKs against global DKs, which
is again in line with our expectation that local DKs can model locality better.

For the comparisons with strong baselines,
an interesting phenomenon worth noting is that though DKs perform better than
Deformable Convolutions on image classification, they fall noticeably short for
object detection measured by mAP.
We speculate that even though both techniques can adapt ERF in theory (as
justified in Section~\ref{sec:analyzing}), directly shifting sampling locations
on data is easier to optimize.
Yet after combining DKs with previous approaches, we can consistently boost performance
for all the methods -- +0.7/+1.2 for Deformable Convolutions on each base
models, and +1.7/+1.1 for Conditional Convolutions.
These findings align with the results from image classification.
We next investigate what DKs learn and why they are compatible with previous
methods in general.

\subsection{What do Deformable Kernels learn?} \label{sec:understanding}

\begin{figure}[t]
    \centering
    \includegraphics[width=0.95\linewidth]{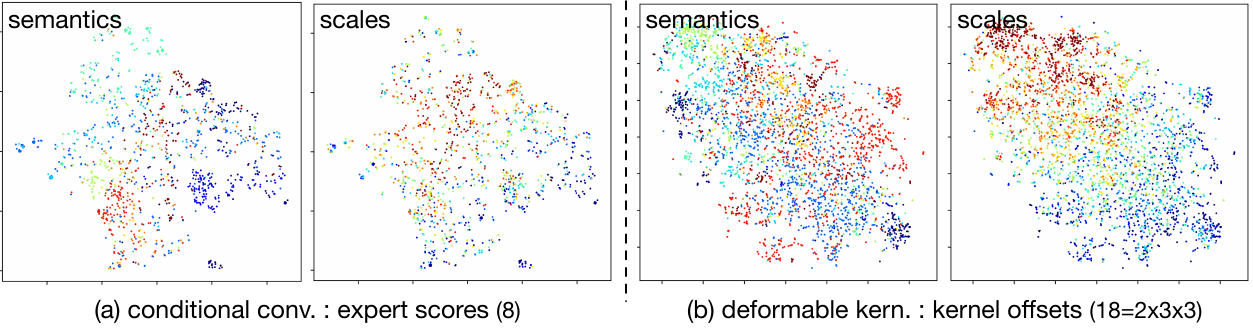}
    \captiont{Semantics vs.\ Scales.}{\
        We show t-SNE results of learned model dynamics using 10 random classes
        of objects from the COCO \texttt{test-dev} split.
        Each point represents an object extracted by ground-truth bounding box,
        whose color either denotes its class label or bounding box
        scale.
        The color of an object scale is its normalized area rank discretized by
        every 10th percentile among all data.
        Numbers inside parentheses indicate the dimension of learned
        dynamics before t-SNE.
        \textbf{(a)}
        The dynamics of Conditional Convolutions are closer to semantics than
        to object scales.
        \textbf{(b)}
        On the contrary, our DKs learn dynamics that are significantly related
        to scales rather than semantics.
    }%
    \vspace{-0.5em}
    \label{fig:sem-vs-scale}
\end{figure}

\begin{figure}[t]
    \centering
    \includegraphics[width=0.95\linewidth]{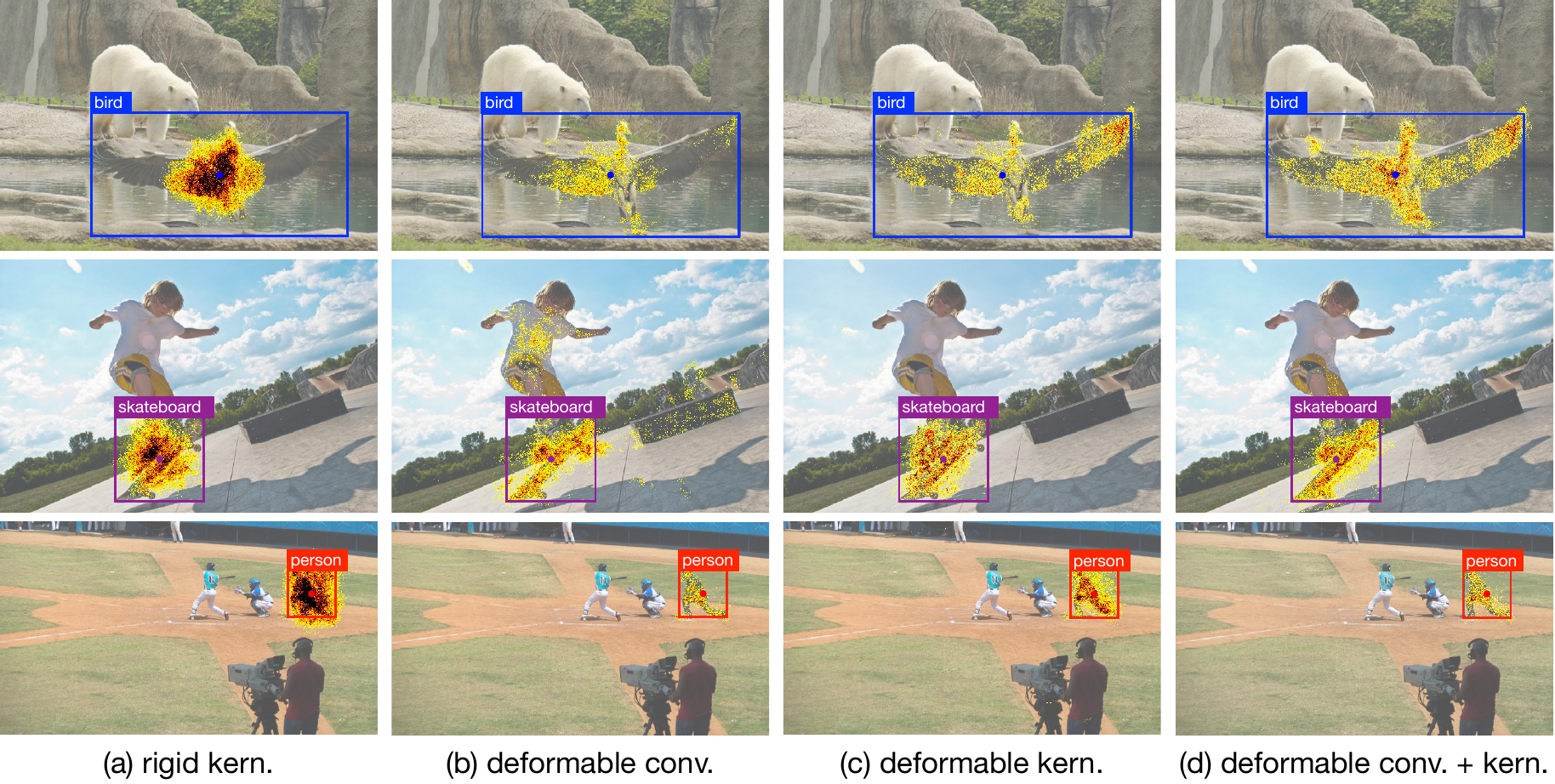}
    \captiont{Learned Effective Receptive Fields.}{\
        We show learned ERFs on three images with \textcolor{large}{large},
        \textcolor{medium}{medium}, and \textcolor{small}{small} objects from
        the COCO \texttt{test-dev} split.
        Given each ground-truth bounding box, we visualize the non-zero ERF
        values of its central point.
        Theoretical RFs cover the whole image for all three examples and we
        thus ignore them in our plots.
        \textbf{(a)} Rigid kernels have strong central effects and a
        Gaussian-like ERF that cannot deal with object deformation alone.
        \textbf{(b)} Deformable Convolutions and \textbf{(c)} Deformable Kernels
        both tune ERFs to data.
        \textbf{(d)} Combining both operators together enables better
        modeling of 2D geometric transformation of objects.
    }%
    \vspace{-0.5em}
    \label{fig:erf_visualization}
\end{figure}

\bfsection{Awareness of Object Scale}
Since deformation is hard to quantify, we use object scale as a rough
proxy to understand what DKs learn.
In Figure~\ref{fig:sem-vs-scale}, we show the
t-SNE~\citep{maaten2008visualizing} of learned model dynamics by the last
convolutional layers in MobileNet-V2 using Conditional Convolution and our DKs.
We validate the finding as claimed by~\citet{yang2019soft} that the experts of
Conditional Convolutions have better correlation with object semantics than
their scales (in reference to Figure~6 from their paper).
Instead, our DKs learn kernel sampling offsets that strongly correlate to
scales rather than semantics.
This sheds light on why the two operators are complementary in our previous
experiments.

\bfsection{Adaptation of Effective Receptive Fields}
To verify our claim that DK indeed adapts ERFs in practice, we show
ERF visualizations on a set of images in which they display different
degrees of deformations.
We compare the results of rigid kernels, Deformable Convolutions, our DKs, and
the combination of the two operators.
For all examples, note that the theoretical receptive field covers every pixel
in the image but ERFs contain only a central portion of it.
Deformable Convolutions and DKs perform similarly in terms of adapting ERFs,
but Deformable Convolutions tend to spread out and have sparse responses while
DKs tend to concentrate and densely activate within an object region.
Combining both operators yields more consistent ERFs that exploit both of their
merits.

\section{Conclusion}

In this paper, we introduced Deformable Kernels~(DKs) to adapt effective
receptive fields~(ERFs) of convolutional networks for object deformation.
We proposed to sample kernel values from the original kernel space.
This in effect samples the ERF in linear networks and also roughly generalizes
to non-linear cases.
We instantiated two variants of DKs and validate our designs, showing
connections to previous works.
Consistent improvements over them and compatibility with them were found, as
illustrated in visualizations.

\bibliography{dk_references}

\begin{thebibliography}{37}
\providecommand{\natexlab}[1]{#1}
\providecommand{\url}[1]{\texttt{#1}}
\expandafter\ifx\csname urlstyle\endcsname\relax
  \providecommand{\doi}[1]{doi: #1}\else
  \providecommand{\doi}{doi: \begingroup \urlstyle{rm}\Url}\fi

\bibitem[Bruna \& Mallat(2013)Bruna and Mallat]{bruna2013invariant}
Joan Bruna and St{\'e}phane Mallat.
\newblock Invariant scattering convolution networks.
\newblock \emph{TPAMI}, 2013.

\bibitem[Cohen \& Welling(2016)Cohen and Welling]{cohen2016group}
Taco Cohen and Max Welling.
\newblock Group equivariant convolutional networks.
\newblock In \emph{ICML}, 2016.

\bibitem[Dai et~al.(2017)Dai, Qi, Xiong, Li, Zhang, Hu, and
  Wei]{dai2017deformable}
Jifeng Dai, Haozhi Qi, Yuwen Xiong, Yi~Li, Guodong Zhang, Han Hu, and Yichen
  Wei.
\newblock Deformable convolutional networks.
\newblock In \emph{CVPR}, 2017.

\bibitem[Deng et~al.(2009)Deng, Dong, Socher, Li, Li, and
  Fei-Fei]{deng2009imagenet}
Jia Deng, Wei Dong, Richard Socher, Li-Jia Li, Kai Li, and Li~Fei-Fei.
\newblock Imagenet: A large-scale hierarchical image database.
\newblock In \emph{CVPR}, 2009.

\bibitem[Esteves et~al.(2018)Esteves, Allen-Blanchette, Zhou, and
  Daniilidis]{polar2018esteves}
Carlos Esteves, Christine Allen-Blanchette, Xiaowei Zhou, and Kostas
  Daniilidis.
\newblock Polar transformer networks.
\newblock In \emph{ICLR}, 2018.

\bibitem[Gibson(1950)]{gibson1950perception}
James~J Gibson.
\newblock The perception of the visual world.
\newblock \emph{Houghton Mifflin}, 1950.

\bibitem[Goyal et~al.(2017)Goyal, Doll{\'a}r, Girshick, Noordhuis, Wesolowski,
  Kyrola, Tulloch, Jia, and He]{goyal2017accurate}
Priya Goyal, Piotr Doll{\'a}r, Ross Girshick, Pieter Noordhuis, Lukasz
  Wesolowski, Aapo Kyrola, Andrew Tulloch, Yangqing Jia, and Kaiming He.
\newblock Accurate, large minibatch sgd: Training imagenet in 1 hour.
\newblock \emph{arXiv preprint arXiv:1706.02677}, 2017.

\bibitem[He et~al.(2016)He, Zhang, Ren, and Sun]{he2016deep}
Kaiming He, Xiangyu Zhang, Shaoqing Ren, and Jian Sun.
\newblock Deep residual learning for image recognition.
\newblock In \emph{CVPR}, 2016.

\bibitem[Howard et~al.(2017)Howard, Zhu, Chen, Kalenichenko, Wang, Weyand,
  Andreetto, and Adam]{howard2017mobilenets}
Andrew~G Howard, Menglong Zhu, Bo~Chen, Dmitry Kalenichenko, Weijun Wang,
  Tobias Weyand, Marco Andreetto, and Hartwig Adam.
\newblock Mobilenets: Efficient convolutional neural networks for mobile vision
  applications.
\newblock In \emph{CVPR}, 2017.

\bibitem[Hu et~al.(2018)Hu, Shen, and Sun]{hu2018squeeze}
Jie Hu, Li~Shen, and Gang Sun.
\newblock Squeeze-and-excitation networks.
\newblock In \emph{CVPR}, 2018.

\bibitem[Hudson \& Manning(2019)Hudson and Manning]{hudson2019learning}
Drew~A Hudson and Christopher~D Manning.
\newblock Learning by abstraction: The neural state machine.
\newblock \emph{arXiv preprint arXiv:1907.03950}, 2019.

\bibitem[Jaderberg et~al.(2015)Jaderberg, Simonyan, Zisserman,
  et~al.]{jaderberg2015spatial}
Max Jaderberg, Karen Simonyan, Andrew Zisserman, et~al.
\newblock Spatial transformer networks.
\newblock In \emph{NeurIPS}, 2015.

\bibitem[Jia et~al.(2016)Jia, De~Brabandere, Tuytelaars, and
  Gool]{jia2016dynamic}
Xu~Jia, Bert De~Brabandere, Tinne Tuytelaars, and Luc~V Gool.
\newblock Dynamic filter networks.
\newblock In \emph{NeurIPS}, 2016.

\bibitem[Kanazawa et~al.(2016)Kanazawa, Sharma, and
  Jacobs]{kanazawa2014locally}
Angjoo Kanazawa, Abhishek Sharma, and David~W. Jacobs.
\newblock Locally scale-invariant convolutional neural networks.
\newblock In \emph{NeurIPS Workshop}, 2016.

\bibitem[Lazebnik et~al.(2006)Lazebnik, Schmid, and Ponce]{lazebnik2006beyond}
Svetlana Lazebnik, Cordelia Schmid, and Jean Ponce.
\newblock Beyond bags of features: Spatial pyramid matching for recognizing
  natural scene categories.
\newblock In \emph{CVPR}, 2006.

\bibitem[Li et~al.(2019)Li, Wang, Hu, and Yang]{li2019selective}
Xiang Li, Wenhai Wang, Xiaolin Hu, and Jian Yang.
\newblock Selective kernel networks.
\newblock In \emph{CVPR}, 2019.

\bibitem[Lin et~al.(2014)Lin, Maire, Belongie, Hays, Perona, Ramanan,
  Doll{\'a}r, and Zitnick]{lin2014microsoft}
Tsung-Yi Lin, Michael Maire, Serge Belongie, James Hays, Pietro Perona, Deva
  Ramanan, Piotr Doll{\'a}r, and C~Lawrence Zitnick.
\newblock Microsoft coco: Common objects in context.
\newblock In \emph{ECCV}, 2014.

\bibitem[Lin et~al.(2017)Lin, Doll{\'a}r, Girshick, He, Hariharan, and
  Belongie]{lin2017feature}
Tsung-Yi Lin, Piotr Doll{\'a}r, Ross Girshick, Kaiming He, Bharath Hariharan,
  and Serge Belongie.
\newblock Feature pyramid networks for object detection.
\newblock In \emph{CVPR}, 2017.

\bibitem[Loshchilov \& Hutter(2017)Loshchilov and Hutter]{loshchilov2016sgdr}
Ilya Loshchilov and Frank Hutter.
\newblock Sgdr: Stochastic gradient descent with warm restarts.
\newblock In \emph{ICLR}, 2017.

\bibitem[Lowe et~al.(1999)]{lowe1999object}
David~G Lowe et~al.
\newblock Object recognition from local scale-invariant features.
\newblock In \emph{ICCV}, 1999.

\bibitem[Luo et~al.(2016)Luo, Li, Urtasun, and Zemel]{luo2016understanding}
Wenjie Luo, Yujia Li, Raquel Urtasun, and Richard Zemel.
\newblock Understanding the effective receptive field in deep convolutional
  neural networks.
\newblock In \emph{NeurIPS}, 2016.

\bibitem[Maaten \& Hinton(2008)Maaten and Hinton]{maaten2008visualizing}
Laurens van~der Maaten and Geoffrey Hinton.
\newblock Visualizing data using t-sne.
\newblock \emph{Journal of machine learning research}, 2008.

\bibitem[Ren et~al.(2015)Ren, He, Girshick, and Sun]{ren2015faster}
Shaoqing Ren, Kaiming He, Ross Girshick, and Jian Sun.
\newblock Faster r-cnn: Towards real-time object detection with region proposal
  networks.
\newblock In \emph{NeurIPS}, 2015.

\bibitem[Rocco et~al.(2017)Rocco, Arandjelovic, and
  Sivic]{rocco2017convolutional}
Ignacio Rocco, Relja Arandjelovic, and Josef Sivic.
\newblock Convolutional neural network architecture for geometric matching.
\newblock In \emph{CVPR}, 2017.

\bibitem[Sandler et~al.(2018)Sandler, Howard, Zhu, Zhmoginov, and
  Chen]{sandler2018mobilenetv2}
Mark Sandler, Andrew Howard, Menglong Zhu, Andrey Zhmoginov, and Liang-Chieh
  Chen.
\newblock Mobilenetv2: Inverted residuals and linear bottlenecks.
\newblock In \emph{CVPR}, 2018.

\bibitem[Shelhamer et~al.(2019)Shelhamer, Wang, and
  Darrell]{shelhamer2019blurring}
Evan Shelhamer, Dequan Wang, and Trevor Darrell.
\newblock Blurring the line between structure and learning to optimize and
  adapt receptive fields.
\newblock \emph{arXiv preprint arXiv:1904.11487}, 2019.

\bibitem[Sifre \& Mallat(2013)Sifre and Mallat]{sifre2013rotation}
Laurent Sifre and St{\'e}phane Mallat.
\newblock Rotation, scaling and deformation invariant scattering for texture
  discrimination.
\newblock In \emph{CVPR}, 2013.

\bibitem[Thomas et~al.(2019)Thomas, Qi, Deschaud, Marcotegui, Goulette, and
  Guibas]{thomas2019kpconv}
Hugues Thomas, Charles~R Qi, Jean-Emmanuel Deschaud, Beatriz Marcotegui,
  Fran{\c{c}}ois Goulette, and Leonidas~J Guibas.
\newblock Kpconv: Flexible and deformable convolution for point clouds.
\newblock In \emph{ICCV}, 2019.

\bibitem[Vaswani et~al.(2017)Vaswani, Shazeer, Parmar, Uszkoreit, Jones, Gomez,
  Kaiser, and Polosukhin]{vaswani2017attention}
Ashish Vaswani, Noam Shazeer, Niki Parmar, Jakob Uszkoreit, Llion Jones,
  Aidan~N Gomez, {\L}ukasz Kaiser, and Illia Polosukhin.
\newblock Attention is all you need.
\newblock In \emph{NeurIPS}, 2017.

\bibitem[Wang et~al.(2019)Wang, Shelhamer, Olshausen, and
  Darrell]{wang2019dynamic}
Dequan Wang, Evan Shelhamer, Bruno Olshausen, and Trevor Darrell.
\newblock Dynamic scale inference by entropy optimization.
\newblock \emph{arXiv preprint arXiv:1908.03182}, 2019.

\bibitem[Wang et~al.(2018)Wang, Girshick, Gupta, and He]{wang2018non}
Xiaolong Wang, Ross Girshick, Abhinav Gupta, and Kaiming He.
\newblock Non-local neural networks.
\newblock In \emph{CVPR}, 2018.

\bibitem[Worrall et~al.(2017)Worrall, Garbin, Turmukhambetov, and
  Brostow]{worrall2017harmonic}
Daniel~E Worrall, Stephan~J Garbin, Daniyar Turmukhambetov, and Gabriel~J
  Brostow.
\newblock Harmonic networks: Deep translation and rotation equivariance.
\newblock In \emph{CVPR}, 2017.

\bibitem[Xie et~al.(2017)Xie, Girshick, Doll{\'a}r, Tu, and
  He]{xie2017aggregated}
Saining Xie, Ross Girshick, Piotr Doll{\'a}r, Zhuowen Tu, and Kaiming He.
\newblock Aggregated residual transformations for deep neural networks.
\newblock In \emph{CVPR}, 2017.

\bibitem[Xiong et~al.(2019)Xiong, Ren, Liao, Wong, and
  Urtasun]{xiong2019deformable}
Yuwen Xiong, Mengye Ren, Renjie Liao, Kelvin Wong, and Raquel Urtasun.
\newblock Deformable filter convolution for point cloud reasoning.
\newblock \emph{arXiv preprint arXiv:1907.13079}, 2019.

\bibitem[Yang et~al.(2019)Yang, Bender, Le, and Ngiam]{yang2019soft}
Brandon Yang, Gabriel Bender, Quoc~V Le, and Jiquan Ngiam.
\newblock Soft conditional computation.
\newblock \emph{arXiv preprint arXiv:1904.04971}, 2019.

\bibitem[Zhang(2019)]{zhang2019making}
Richard Zhang.
\newblock Making convolutional networks shift-invariant again.
\newblock In \emph{ICML}, 2019.

\bibitem[Zhu et~al.(2019)Zhu, Hu, Lin, and Dai]{zhu2019deformable}
Xizhou Zhu, Han Hu, Stephen Lin, and Jifeng Dai.
\newblock Deformable convnets v2: More deformable, better results.
\newblock In \emph{CVPR}, 2019.

\end{thebibliography}
\bibliographystyle{iclr2020_conference}

\clearpage
\appendix
\section{Computation Flow of Deformable Kernels}

We now cover more details on implementing DKs by elaborating the computation
flow of their forward and backward passes.
We will focus on the local DK given its superior performance in practice.
The extension to global DK implementation is straight-forward.

\subsection{Forward Pass}
\begin{figure}[t]
    \centering
    \includegraphics[width=0.8\linewidth]{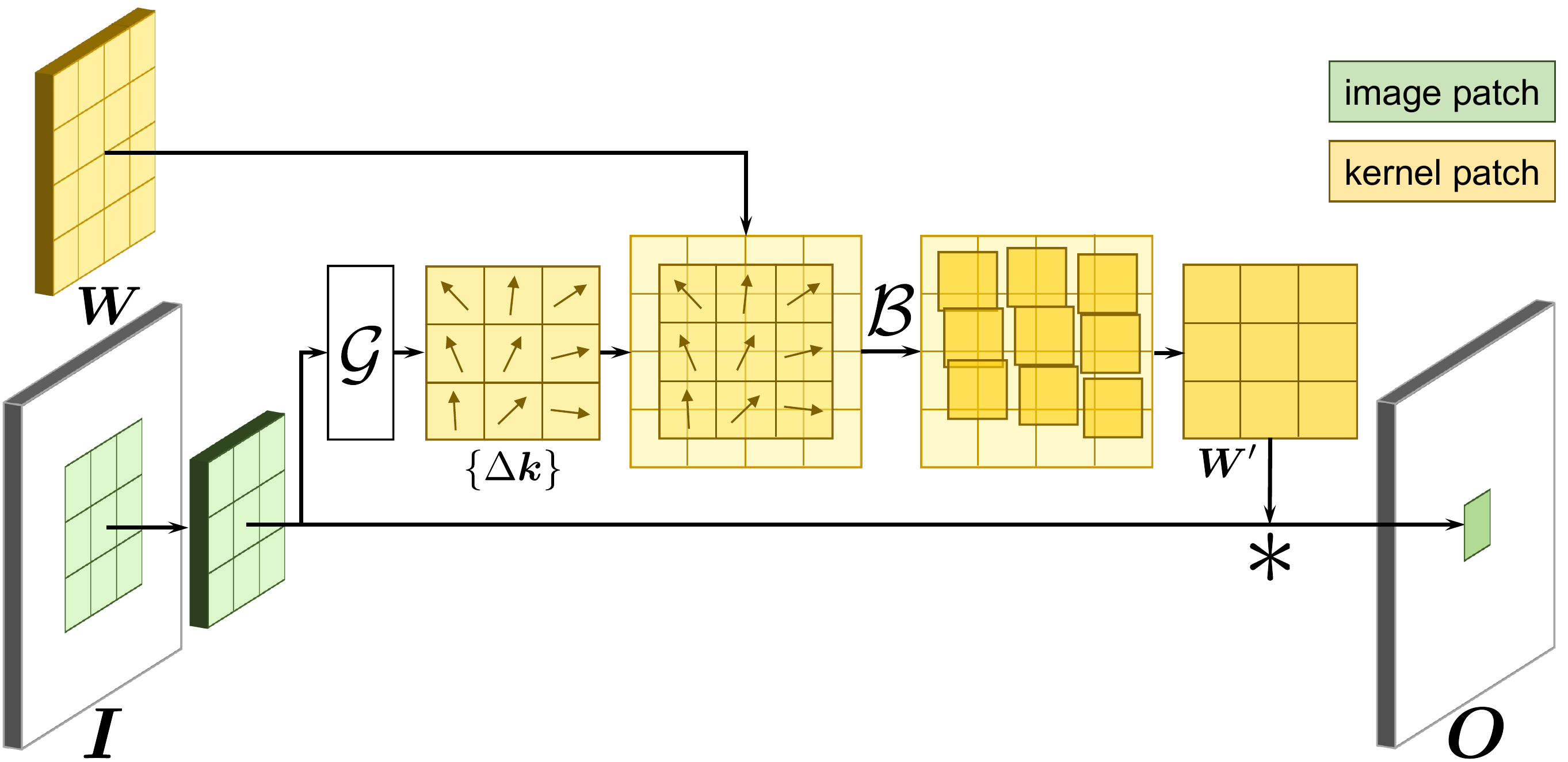}
    \captiont{
            Illustration of feed-forwarding through a 3$\times$3 local
            Deformable Kernel from a 4$\times$4 scope.
        }{\
        For each input patch, local DK first generates a group of kernel
        offsets $\{\Delta \vk\}$ from input feature patch using the light-weight
        generator $\gG$ (a 3$\times$3 convolution of rigid kernel).
        Given the original kernel weights $\mW$ and the offset group $\{\Delta \vk\}$, DK
        samples a new set of kernel $\mW'$ using a bilinear sampler $\gB$.
        Finally, DK convolves the input feature map and the sampled kernels to complete the
        whole computation.
    }%
    \vspace{-0.5em}
    \label{fig:local-dk}
\end{figure}

In Section~\ref{sec:dk}, we introduce a kernel offset
generator $\gG$ and a bilinear sampler $\gB$.
Figure~\ref{fig:local-dk} illustrates an example of the forward pass.

Concretely, given a kernel $\mW$ and a learned group of kernel offsets $\{\Delta \vk\}$ on
top of a regular 2D grid $\{\vk\}$, we can resample a new kernel $\mW'$ by
a bilinear operator $\gB$ as
\begin{align}
\small
    \mW'
    &\equiv
    \mW_{\vk + \Delta \vk}
    =
    \sum_{\vk' \in \gK}
    \gB(\vk + \Delta \vk, \vk')
    \mW_{\vk'}
    ,\\
    \text{where  }
    \gB(\vk + \Delta \vk, \vk')
    &=
    \max(0, 1 - |k_x + \Delta k_x -k'_x|)
    \cdot
    \max(0, 1 - |k_y + \Delta k_y-k'_y|). \nonumber
\end{align}
Given this resampled kernel, DK convolves it with the input image just as in normal
convolutions using rigid kernels, characterized by Equation~\ref{eq:conv2d}.

\subsection{Backward Pass}
The backward pass of local DK consists of three types of gradients: (1)
the gradient to the data of the previous layer, (2) the gradient to the full scope
kernel of the current layer and (3) the additional gradient to the kernel
offset generator of the current layer.
The first two types of gradients share same forms of the computation comparing
to the normal convolutions.
We now cover the computation for the third flow of gradient that directs where
to sample kernel values.

In the context of Equation~\ref{eq:dk2d}, the partial derivative of a output
item $\mO_{\vj}$ w.r.t.  $x$ component of a given kernel offset $\Delta k_x$ (similar for its $y$ component $\Delta k_y$) can be computed as
\begin{align}
    \frac{\partial \mO_{\vj}}{\partial \Delta k_x}
    &=
    \sum_{\vk}
    \mI_{\vj + \vk}
    \bigg(
        \sum_{\vk'}
        \mW_{\vk'}
        \frac{\partial \gB (\vk + \Delta \vk, \vk')}{\partial \Delta k_x}
    \bigg), \\
    \text{where  }
    \frac{\partial \gB (\vk + \Delta \vk, \vk')}{\partial \Delta k_x}
    &=
    \max(0, 1 - |k_y + \Delta k_y-k'_y|)
    \cdot
    \begin{cases}
    0 & |k_x + \Delta k_x-k'_x| \geq 1 \\
    1 & k_x + \Delta k_x < k'_x\\
    -1 & k_x + \Delta k_x \geq k'_x\\
    \end{cases}. \nonumber
\end{align}

\section{Network Architectures}

Table~\ref{tab:r50dw-arch} shows the comparison between the original
ResNet-50~\citep{he2016deep} and our modified ResNet-50-DW.
The motivation of introducing depthwise convolutions to ResNet is to accelerate
the computation of local DKs based on our current implementations.
The ResNet-50-DW model has similar model capacity/complexity and performance
(see Table~\ref{tab:cls_ablation_space}) compared to its non-depthwise
counterpart, making it an ideal base architecture
for our experiments.

On the other hand, in all of our experiments,
MobileNet-V2~\citep{sandler2018mobilenetv2} base model is left untouched.

\begin{table}
    \centering
    \renewcommand\arraystretch{1.1}
    \newcommand{\tabincell}[2]{\begin{tabular}{@{}#1@{}}#2\end{tabular}}

    \scalebox{1}{
        \begin{tabular}{c|c|c}
            \Xhline{1.0pt}
            Output & ResNet-50 & ResNet-50-DW \\
            \Xhline{0.7pt}
            112 $\times$ 112 &
            \multicolumn{2}{c}{7 $\times$ 7, 64, stride 2} \\
            \hline
            56 $\times$ 56 &
            \multicolumn{2}{c}{3 $\times$ 3 max pool, stride 2} \\
            \hline
            56 $\times$ 56
            &
            $
            \begin{bmatrix}
                \begin{array}{l}
                    $1 $ \times $ 1$ , $ 64$ \\
                    $3 $ \times $ 3$ , $ 64$ \\
                    $1 $ \times $ 1$ , $ 256$
                \end{array}
            \end{bmatrix} \times$ 3$
            $
            &
            $
            \begin{bmatrix}
                \begin{array}{l}
                    $1 $ \times $ 1$ , $ 128$ \\
                    $3 $ \times $ 3$ , $ 128$, G = $ 128$ \\
                    $1 $ \times $ 1$ , $ 256$
                \end{array}
            \end{bmatrix} \times$ 3$
            $
            \\
            \hline
            28 $\times$ 28 &
            $
            \begin{bmatrix}
                \begin{array}{l}
                    $1 $ \times $ 1$, $ 128$ \\
                    $3 $ \times $ 3$, $ 128$ \\
                    $1 $ \times $ 1$, $ 512$
                \end{array}
            \end{bmatrix} \times$ 4$
            $
            &
            $
            \begin{bmatrix}
               \begin{array}{l}
                   $1 $ \times $ 1$, $ 256$ \\
                   $3 $ \times $ 3$, $ 256$, G = $ 256$ \\
                   $1 $ \times $ 1$, $ 512$
               \end{array}
            \end{bmatrix} \times$ 4$
            $
            \\
            \hline
            14 $\times$ 14
            &
            $
            \begin{bmatrix}
               \begin{array}{l}
                   $1 $\times$ 1$, $ 256$ \\
                   $3 $\times$ 3$, $ 256$ \\
                   $1 $\times$ 1$, $ 1024$
               \end{array}
            \end{bmatrix} \times$ 6$
            $
            &
            $
            \begin{bmatrix}
               \begin{array}{l}
                   $1 $\times$ 1$, $ 512$  \\
                   $3 $\times$ 3$, $ 512$, G = $ 512$ \\
                   $1 $\times$ 1$, $ 1024$
               \end{array}
            \end{bmatrix} \times$ 6$
            $
            \\
            \hline
            7 $\times$ 7
            &
            $
            \begin{bmatrix}
               \begin{array}{l}
                   $1 $\times$ 1$, $ 512$ \\
                   $3 $\times$ 3$, $ 512$ \\
                   $1 $\times$ 1$, $ 2048$
               \end{array}
            \end{bmatrix} \times$ 3$
            $
            &
            $
            \begin{bmatrix}
               \begin{array}{l}
                   $1 $\times$ 1$, $ 1024$  \\
                   $3 $\times$ 3$, $ 1024$, G = $1024$  \\
                   $1 $\times$ 1$, $ 2048$
               \end{array}
            \end{bmatrix} \times$ 3$
            $
            \\
            \hline
            1 $\times$ 1 &
            \multicolumn{2}{c} {7 $\times$ 7 global average pool, 1000-d $fc$, {softmax}} \\
            \Xhline{0.7pt}
            \#P (M) &
            25.6 &
            23.7
            \\
            \hline
            GFLOPs &
            3.86 &
            3.82
            \\
            \Xhline{1.0pt}
    \end{tabular}}
    \vspace{+4pt}
    \captiont{
        Network architecture of our ResNet-50-DW comparing to the original
        ResNet-50
    }{
        Inside the brackets are the general shape of a residual block,
        including filter sizes and feature dimensionalities.
        The number of stacked blocks on each stage is presented outside the
        brackets.
        ``$G=$~128'' suggests the depthwise convolution with 128 input
        channels.
        Two models have similar numbers of parameters and FLOPs.
        At the same time, depthwise convolutions facilitate the computation
        efficiency of our Deformable Kernels.
    }
    \label{tab:r50dw-arch}
    \vspace{-2pt}
\end{table}

\section{Additional Comparison of Effective Receptive Fields}
\begin{figure}[t]
    \centering
    \includegraphics[width=0.8\linewidth]{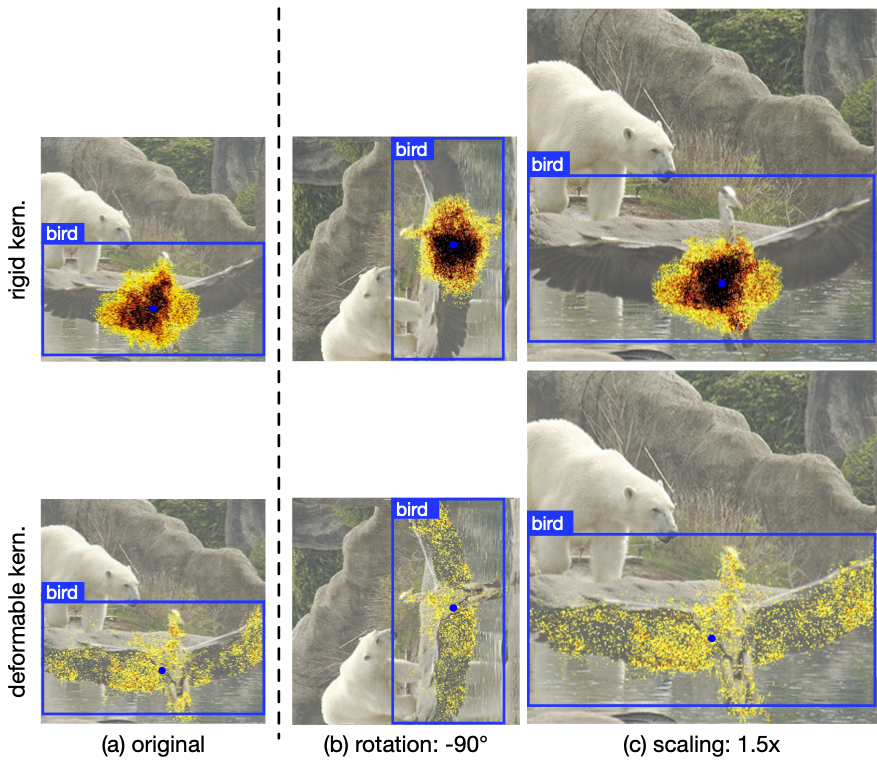}
    \captiont{
            Effective Receptive Field Comparison between rigid kernels and DKs
            under different kinds of Object Deformation.
        }{\
        At each row and from left to right, we show the original image
        (1300$\times$800), the image rotated by -90 degrees and the image scaled
        by 1.5 times.  Images are cropped and resized for the typesetting
        purpose.
    }%
    \vspace{-0.5em}
    \label{fig:additional_erf_visualization}
\end{figure}

We here show additional comparison of ERFs when objects have different kinds
of deformations in Figure~\ref{fig:additional_erf_visualization}.
Comparing to baseline, our method can adapt ERFs to be more persistent to
object’s semantic rather than its geometric configuration.

\section{Additional Experiment Details}
\bfsection{Image Classification}
Similar to~\citet{goyal2017accurate,loshchilov2016sgdr}, training is
performed by SGD for 90 epochs with momentum 0.9 and batch size 256.
We set our learning rate of~$10^{-1}$ so that it linearly warms up from zero within
first 5 epochs.
A cosine training schedule is applied over the training epochs.
We use scale and aspect ratio augmentation with color perturbation as
standard data augmentations.
We evaluate the performance of trained models on the ImageNet 2012 validation
set.
The images are resized so that the shorter side is of 256 pixels.
We then centrally crop~$224 \times 224$ windows from the images as input to
measure recognition accuracy.

\end{document}